\documentclass{article}
\usepackage{color}
\usepackage{url}
\usepackage{spconf,graphicx}
\usepackage{amssymb}

\newcommand{\ignore}[1]{}

\def\x{\mathbf x}

\title{Deep Residual Learning for Small-Footprint Keyword Spotting}

\name{Raphael Tang \qquad Jimmy Lin}
\address{David R. Cheriton School of Computer Science\\
	University of Waterloo\\
	\texttt{\{r33tang,jimmylin\}@uwaterloo.ca}}

\begin{document}

\maketitle

\begin{abstract}
We explore the application of deep residual learning and dilated
convolutions to the keyword spotting task, using the recently-released
Google Speech Commands Dataset as our benchmark. Our best residual
network (ResNet) implementation significantly outperforms Google's
previous convolutional neural networks in terms of accuracy. By
varying model depth and width, we can achieve compact models that also
outperform previous small-footprint variants. To our knowledge, we are
the first to examine these approaches for keyword spotting, and our
results establish an open-source state-of-the-art reference to support
the development of future speech-based interfaces.
\end{abstract}


\section{Introduction}
\label{sec:intro}

The goal of keyword spotting is to detect a relatively small set of
predefined keywords in a stream of user utterances, usually in the
context of an intelligent agent on
a mobile phone or a consumer ``smart home'' device. Such a
capability complements full automatic speech recognition, which is
typically performed in the cloud. Because cloud-based interpretation
of speech input requires transferring audio recordings from the user's
device, there are significant privacy implications. Therefore,
on-device keyword spotting has two main uses:\ First, recognition of
common commands such as ``on'' and ``off'' as well as other frequent
words such as ``yes'' and ``no'' can be accomplished directly on the
user's device, thereby sidestepping any potential privacy
concerns. Second, keyword spotting can be used to detect ``command
triggers'' such as ``hey Siri'', which provide explicit cues for
interactions directed at the device.
It is additionally desirable that such models have a small footprint
(for example, measured in the number of model parameters) so
they can be deployed on low power and performance-limited
devices.

In recent years, neural networks have been shown to provide effective
solutions to the small-footprint keyword spotting problem. Research
typically focuses on a tradeoff between achieving high detection
accuracy and having a small footprint. Compact models are usually
variants derived from a full model that sacrifice accuracy for
a smaller model footprint, often via some form of sparsification.

In this work, we focus on convolutional neural networks (CNNs), a
class of models that has been successfully applied to small-footprint
keyword spotting in recent years. In particular, we explore the use of
residual learning techniques and dilated convolutions. On the
recently-released Google Speech Commands Dataset, which provides a
common benchmark for keyword spotting, our full residual network model
outperforms Google's previously-best CNN~\cite{keywordcnn}
(95.8\% vs.\ 91.7\% in accuracy). We can tune the depth
and width of our networks to target a desired tradeoff between model
footprint and accuracy:\ one variant is able to achieve accuracy only
slightly below Google's best CNN with a 50$\times$ reduction in model
parameters and an 18$\times$ reduction in the number of multiplies in
a feedforward inference pass. This model far outperforms previous compact CNN
variants.

\section{Related Work}
\label{sec:rel_work}

Deep residual networks (ResNets)~\cite{resnet} represent a groundbreaking
advance in deep learning that has allowed researchers to successfully
train deeper networks. They were first applied to image
recognition, where they contributed to a significant
jump in state-of-the-art performance~\cite{resnet}. ResNets have subsequently been
applied to speaker identification~\cite{resnetsv} and automatic speech
recognition~\cite{resnetasr, resnetasr2}. This paper explores the
application of deep residual learning techniques to the keyword spotting task.

The application of neural networks to keyword spotting, of
course, is not new. Chen et al.~\cite{keyworddnn} applied a standard
multi-layer perceptron to achieve significant improvements over
previous HMM-based approaches. Sainath and Parada~\cite{keywordcnn} built on
that work and achieved better results using convolutional neural
networks (CNNs). They specifically cited reduced model footprints (for
low-power applications) as a major motivation in moving to CNNs.

Despite more recent work in applying recurrent neural networks to the
keyword spotting task~\cite{keywordrnn,SunMing_etal_2017}, we
focus on the family of CNN models for several reasons. CNNs
today remain the standard baseline for small-footprint keyword
spotting---they have a straightforward architecture, are relatively
easy to tune, and have implementations in multiple deep learning
frameworks (at least TensorFlow~\cite{dataset} and
PyTorch~\cite{honk}). We are not aware of any publicly-available
implementations of recurrent architectures to compare against. We
believe that residual learning techniques form a yet
unexplored direction for the keyword spotting task, and that our use
of dilated convolutions achieves the same goal that proponents
of recurrent architectures tout, the ability to capture long(er)-range
dependencies.

\section{Model Implementation}
\label{sec:impl}

This section describes our base model and its variants. All code
necessary to replicate our experiments has been made open source in
our GitHub repository.\footnote{https://github.com/castorini/honk/}

\subsection{Feature Extraction and Input Preprocessing}

For feature extraction, we first apply a band-pass filter of 20Hz/4kHz
to the input audio to reduce noise. Forty-dimensional Mel-Frequency
Cepstrum Coefficient (MFCC) frames are then constructed and stacked
using a 30ms window and a 10ms frame shift. All frames are stacked
across a 1s interval to form the two-dimensional input to our models.

\subsection{Model Architecture}

\begin{figure}
	\centering
	\includegraphics[width=0.49\textwidth]{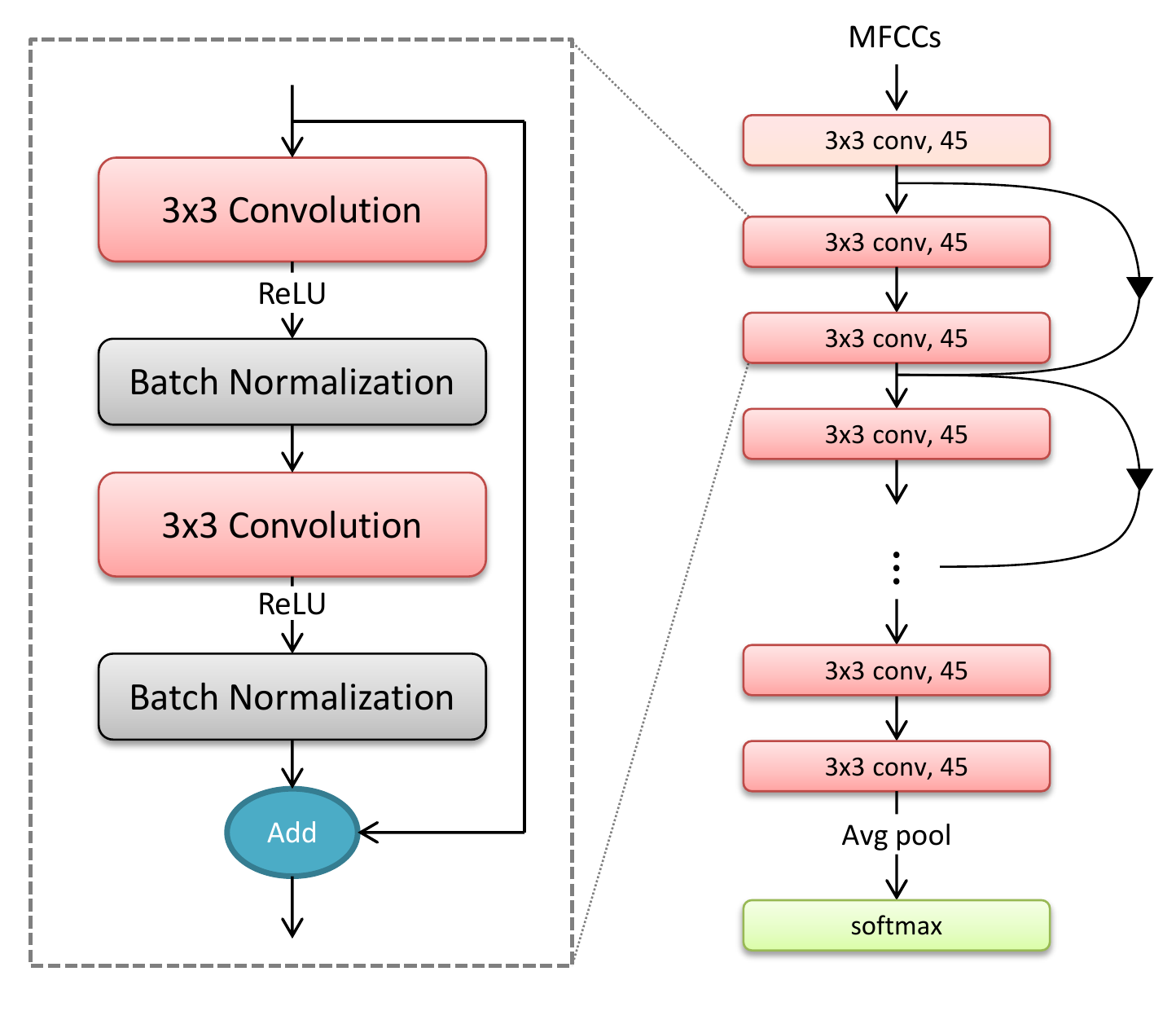}
        \vspace{-0.5cm}
	\caption{Our full architecture, with a magnified residual block.}
	\label{fig:full_arch_res}
\end{figure}

Our architecture is similar to that of He et al.~\cite{resnet}, who postulated that it may be 
easier to learn residuals than to learn the original mapping for deep
convolutional neural networks. They found that additional 
layers in deep networks cannot be merely ``tacked on'' to shallower nets.
Specifically, He et al.~proposed that it may be easier to learn the 
residual $H(\x) = F(\x) + \x$ instead of the true mapping 
$F(\x)$, since it is empirically difficult to learn the identity mapping 
for $F$ when the model has unnecessary depth. In residual networks (ResNets),
residuals are 
expressed via connections between layers (see Figure \ref{fig:full_arch_res}), where an 
input $\x$ to layer $i$ is added to the output of some downstream layer $i + k$,
enforcing the residual definition $H(\x) = F(\x) + \x$.

Following standard ResNet architectures, our residual block begins with a 
bias-free convolution layer with weights $\mathbf{W} \in \mathbb{R}^{(m \times 
r) \times n}$, where $m$ and $r$ are the width and height, respectively, and 
$n$ the number of feature maps. After the convolution layer, there are ReLU 
activation units and---instead of dropout---a batch normalization~\cite{bn} 
layer. In addition to using residual blocks, we also use a $(d_w, d_h)$ 
convolution dilation~\cite{dilated_conv} to increase the receptive field of the 
network, which allows us to consider the one-second input in its entirety using 
a smaller number of layers. To expand our input for the residual blocks, which 
requires inputs and outputs of equal size throughout, our entire architecture 
starts with a convolution layer with weights $\mathbf{W} \in \mathbb{R}^{(m 
\times r) \times n}$. A separate non-residual convolution layer and batch 
normalization layer are further appended to the chain of residual blocks, as 
shown in Figure~\ref{fig:full_arch_res} and Table~\ref{table:full_arch}.

\begin{figure}
	\centering
	\includegraphics[width=0.48\textwidth]{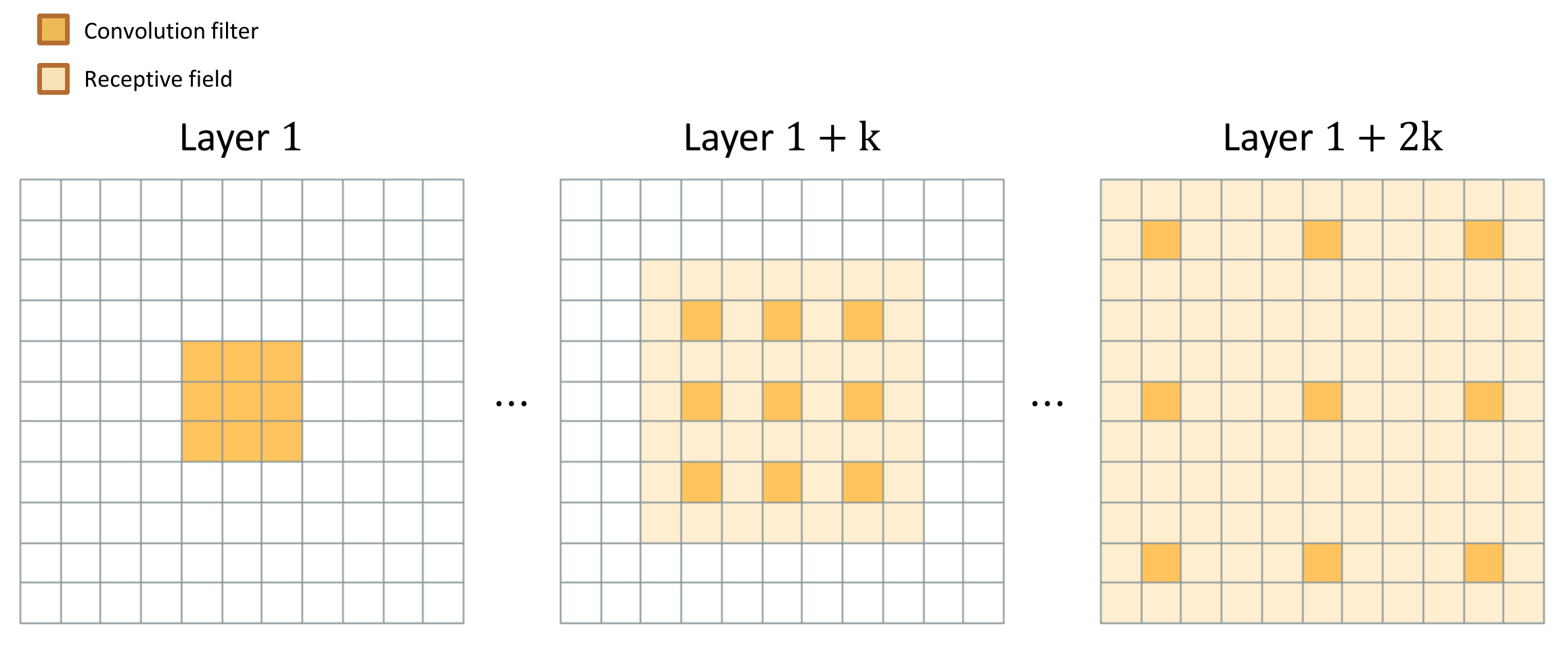}
        \vspace{-0.75cm}
	\caption{Exponentially increasing dilated convolutions; in this case, $k = 
	1$.}
	\label{fig:dilated_conv}
        \vspace{0.25cm}
\end{figure}

\begin{table}
	\begin{center}
		\begin{tabular}{ r | c c c c c | c c}
			\hline
			type & $m$ & $r$ & $n$ & $d_w$ & $d_h$ & Par. & Mult.\\
			\hline
			conv & 3 & 3 & 45 & - & - & 405 & 1.52M\\
			res $\times$ 6 & 3 & 3 & 45 & $2^{\lfloor\frac{i}{3}\rfloor}$ & 
			$2^{\lfloor\frac{i}{3}\rfloor}$ & 219K & 824M\\
			conv & 3 & 3 & 45 & 16 & 16 & 18.2K & 68.6M\\
			bn & - & - & 45 & - & - & - & 169K\\
			avg-pool & - & - & 45 & - & - & - & 45\\
			softmax & - & - & 12 & - & - & 540 & 540\\
			\hline
			\hline
			Total & - & - & - & - & - & 238K & 894M\\
			\hline
		\end{tabular}
	\end{center}
	\vspace{-0.45cm}
	\caption{Parameters used for \texttt{res15}, along with the number of 
	parameters and multiplies.}
	\label{table:full_arch}
	\vspace{0.1cm}
\end{table}

Our base model, which we refer to as \texttt{res15},
comprises six such residual blocks and $n = 45$ feature maps (see Figure 
\ref{fig:full_arch_res}).
For dilation, as illustrated in Figure \ref{fig:dilated_conv}, an exponential sizing 
schedule~\cite{dilated_conv} is used:\ at layer $i$, the dilation is $d_w = d_h 
= 2^{\lfloor\frac{i}{3}\rfloor}$, resulting in a total receptive field of $125 
\times 125$. As is standard in ResNet architectures, all output is zero-padded 
at each layer and finally 
average-pooled and fed into a fully-connected softmax layer.
Following previous work, we measure the ``footprint'' of a model in terms of
two quantities:\ the number of parameters in the model and the number
of multiplies that are required for a full feedforward inference pass.
Our architecture uses roughly 238K parameters and 894M multiplies
(see Table~\ref{table:full_arch} for the exact breakdown).

\begin{table}
	\begin{center}
		\begin{tabular}{ r | c c c | c c}
			\hline
			type & $m$ & $r$ & $n$ & Par. & Mult.\\
			\hline
			conv & 3 & 3 & 19 & 171 & 643K\\
			avg-pool & 4 & 3 & 19 & - & 6.18K\\
			res $\times$ 3 & 3 & 3 & 19 & 19.5K & 5.0M\\
			avg-pool & - & - & 19 & - & 19\\
			softmax & - & - & 12 & 228 & 228\\
			\hline
			\hline
			Total & - & - & - & 19.9K & 5.65M\\
			\hline
		\end{tabular}
	\end{center}
	\vspace{-0.45cm}
	\caption{Parameters used for \texttt{res8-narrow}.}
	\label{table:compact_arch}
	\vspace{0.1cm}
\end{table}

\begin{table}
	\begin{center}
		\begin{tabular}{ r | c c c | c c }
			\hline
			type & $m$ & $r$ & $n$ & Par. & Mult.\\
			\hline
			conv & 3 & 3 & 45 & 405 & 1.80M\\
			avg-pool & 2 & 2 & 45 & - & 45K\\
			res $\times$ 12 & 3 & 3 & 45 & 437K & 378M\\
			avg-pool & - & - & 45 & - & 45\\
			softmax & - & - & 12 & 540 & 540\\
			\hline
			\hline
			Total & - & - & - & 438K & 380M\\
			\hline
		\end{tabular}
	\end{center}
	\vspace{-0.45cm}
	\caption{Parameters used for \texttt{res26}.}
	\label{table:deep_arch}
	\vspace{0.1cm}
\end{table}

To derive a compact small-footprint model, one simple approach is to
reduce the depth of the network. We tried cutting the number of
residual blocks in half to three, yielding a model we call {\tt res8}.
Because the footprint of {\tt res15} arises from its
width as well as its depth, the compact model adds a $4 \times 3$ average-pooling layer
after the first convolutional layer, reducing the size of the time and
frequency dimensions by a factor of four and three,
respectively. Since the average pooling layer sufficiently reduces the
input dimension, we did not use dilated convolutions in this variant.

In the opposite direction, we explored the effects of deeper models.
We constructed a model with double the number of residual blocks (12)
with 26 layers, which we refer to as \texttt{res26}. To make
training tractable, we prepend a $2\times 2$ average-pooling layer to
the chain of residual blocks. Dilation is also not used, since the
receptive field of 25 3$\times$3 convolution filters is large enough
to cover our input size.

In addition to depth, we also varied model width. All
models described above used $n = 45$ feature maps, but we also considered
variants with $n = 19$ feature maps, denoted by \texttt{-narrow}
appended to the base model's name. A detailed breakdown of the
footprint of \texttt{res8-narrow}, our best compact model, is shown in
Table~\ref{table:compact_arch}; the same analysis for our deepest and
widest model, \texttt{res26}, is shown in Table~\ref{table:deep_arch}.

\section{Evaluation}

\subsection{Experimental Setup}

We evaluated our models using Google's Speech Commands
Dataset~\cite{dataset}, which was released in August 2017 under a
Creative Commons license.\footnote{\url{https://research.googleblog.com/2017/08/launching-speech-commands-dataset.html}}
The dataset contains 65,000 one-second long
utterances of 30 short words by thousands of different people, as well
as background noise samples such as pink noise, white noise, and human-made
sounds. The blog post announcing the data release also references
Google's TensorFlow implementation of Sainath and Parada's models,
which provide the basis of our comparisons.

Following Google's implementation, our task is to discriminate among 12
classes:\ ``yes,'' ``no,'' ``up,'' ``down,'' ``left,'' ``right,''
``on,'' ``off,'' ``stop,'' ``go'', unknown, or silence. Our
experiments followed exactly the same procedure as the TensorFlow reference.
The Speech Commands Dataset was split into training,
validation, and test sets, with 80\% training, 10\% validation,
and 10\% test. This results in roughly 22,000 examples for
training and 2,700 each for validation and testing. For consistency
across runs, the SHA1-hashed name of the audio file from the dataset
determines the split.

To generate training data, we followed Google's preprocessing
procedure by adding background noise to each sample with a probability
of $0.8$ at every epoch, where the noise is chosen randomly from the
background noises provided in the dataset. Our
implementation also performs a random time-shift of $Y$ milliseconds
before transforming the audio into MFCCs, where
$Y\sim\textsc{Uniform}[-100, 100]$. In order to accelerate the
training process, all preprocessed inputs are cached for reuse across
different training epochs. At each epoch, 30\% of the cache is evicted.

Accuracy is our main metric of quality, which is simply measured as
the fraction of classification decisions that are correct. For each instance,
the model outputs its most likely prediction, and is not given the option of ``don't know''.
We also plot receiver operating characteristic (ROC) curves, where the $x$
and $y$ axes show false alarm rate (FAR) and false reject
rate (FRR), respectively. For a given sensitivity threshold---defined
as the minimum probability at which a class is considered
positive during evaluation---FAR and FRR represent the probabilities
of obtaining false positives and false negatives, respectively. By
sweeping the sensitivity interval $[0.0, 1.0]$, curves for each of the
keywords are computed and then averaged vertically to produce the
overall curve for a particular model. Curves with less area under the
curve (AUC) are better.

\subsection{Model Training}

Mirroring the ResNet paper~\cite{resnet}, we used stochastic gradient 
descent with a momentum of 0.9 and a starting learning rate of 0.1, which 
is multiplied by 0.1 on plateaus. We also experimented with Nesterov momentum, 
but we found slightly decreased learning performance in terms of cross entropy 
loss and test accuracy. We used a mini-batch size of 64 and $L_2$ weight decay 
of $10^{-5}$. Our models were trained for a total of 26 epochs, resulting in 
roughly 9,000 training steps.

\subsection{Results}

\begin{table}[t]
	\begin{center}
		\begin{tabular}{ l c | c c} 
			\hline
			Model & Test accuracy & Par. & Mult.\\
			\hline
			\texttt{trad-fpool3} & 90.5\% $\pm$ 0.297 & 1.37M & 125M\\
			\texttt{tpool2} & 91.7\% $\pm$ 0.344 & 1.09M & 103M\\
			\texttt{one-stride1} & 77.9\% $\pm$ 0.715 & 954K & 5.76M\\
			\hline
			\texttt{res15}        & 95.8\% $\pm$ 0.484 & 238K & 894M\\
			\texttt{res15-narrow}        & 94.0\% $\pm$ 0.516 & 42.6K & 160M\\
			\hline
			\texttt{res26} & 95.2\% $\pm$ 0.184 & 438K & 380M\\
			\texttt{res26-narrow}        & 93.3\% $\pm$ 0.377 & 78.4K & 68.5M\\
			\hline
			\texttt{res8} & 94.1\% $\pm$ 0.351 & 110K & 30M\\
			\texttt{res8-narrow} & 90.1\% $\pm$ 0.976 & 19.9K & 5.65M\\
			\hline
		\end{tabular}
	\end{center}
	\vspace{-0.45cm}
	\caption{Test accuracy of each model with 95\% confidence
          intervals (across five trials), as well as footprint size in
          terms of number of parameters and multiplies.}
	\label{table:results}
	\vspace{-0.2cm}
\end{table}

Since our own networks are implemented in PyTorch, we used our PyTorch
reimplementations of Sainath and Parada's models as a point of
comparison. We have previously confirmed that our PyTorch
implementation achieves the same accuracy as the original TensorFlow
reference~\cite{honk}. Our ResNet models are compared against three CNN variants
proposed by Sainath and Parada:\ \texttt{trad-fpool3}, which is their
base model; \texttt{tpool2}, the most accurate variant of those they
explored; and \texttt{one-stride1}, their best compact variant. The
accuracies of these models are shown in
Table~\ref{table:results}, which also shows the 95\% confidence intervals
from five different optimization trials with different random
seeds. The table provides the number of model parameters as well as
the number of multiplies in an inference pass. We see that
\texttt{tpool2} is indeed the best performing model, slightly better
than \texttt{trad-fpool3}. The \texttt{one-stride1} model
substantially reduces the model footprint, but this comes at a steep price
in terms of accuracy.

The performance of our ResNet variants is also shown in
Table~\ref{table:results}. Our base \texttt{res15} model achieves
significantly better accuracy than any of the previous Google CNNs
(the confidence intervals do not overlap). This model requires fewer
parameters, but more multiplies, however. The ``narrow'' variant of
\texttt{res15} with fewer feature maps sacrifices accuracy, but
remains significantly better than the Google CNNs (although it still
uses $\sim$30\% more multiplies).

Looking at our compact \texttt{res8} architecture, we see that the ``wide''
version strictly dominates all the Google models---it achieves
significantly better accuracy with a smaller footprint. The ``narrow''
variant reduces the footprint even more, albeit with a small degradation in
performance compared to \texttt{tpool2}, but requires 50$\times$ fewer
model parameters and 18$\times$ fewer multiplies. Both models are
far superior to Google's compact variant, \texttt{one-stride1}.

\begin{figure}
\vspace{-0.4cm}
\centering
\includegraphics[width=0.49\textwidth]{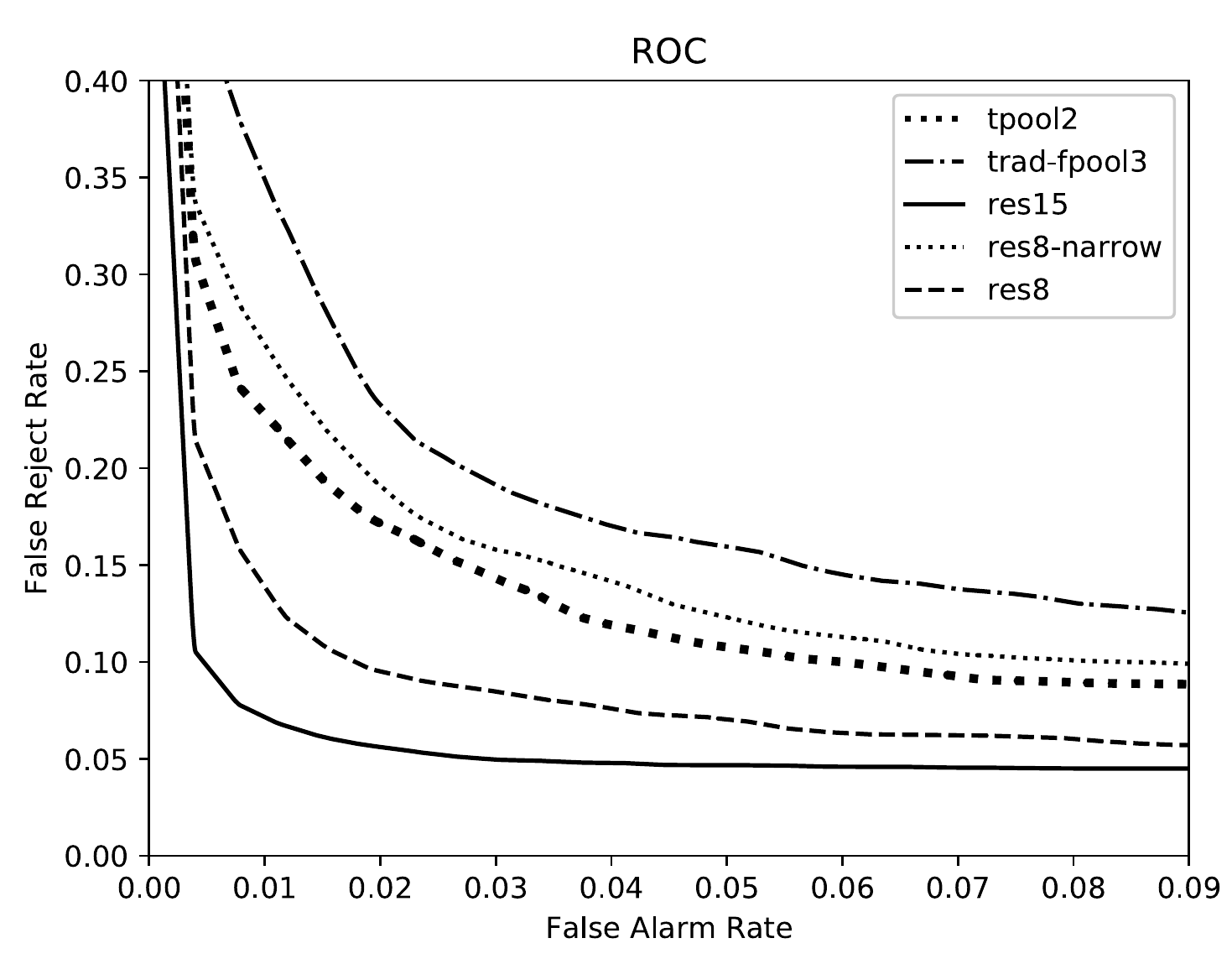}
\vspace{-0.7cm}
\caption{ROC curves for different models.}
\label{fig:roc}
\vspace{-0.2cm}
\end{figure}

Turning our attention to the deeper variants, we see that
\texttt{res26} has lower accuracy than \texttt{res15}, suggesting that
we have overstepped the network depth for which we can properly
optimize model parameters. Comparing the narrow vs.\ wide variants
overall, it appears that width (the number of feature maps)
has a larger impact on accuracy than depth.

We plot the ROC curves of selected models in Figure~\ref{fig:roc},
comparing the two competitive baselines to \texttt{res8},
\texttt{res8-narrow}, and \texttt{res15}. The remaining models were
less interesting and thus omitted for clarity. These curves are
consistent with the accuracy results presented in
Table~\ref{table:results}, and we see that \texttt{res15} dominates
the other models in performance at all operating points.

\section{Conclusions and Future Work}

This paper describes the application of deep residual learning
and dilated convolutions to the keyword spotting problem. Our work is
enabled by the recent release of Google's Speech Commands
Dataset, which provides a common benchmark for this task. Previously,
related work was mostly incomparable because papers relied on
private datasets. Our work establishes new,
state-of-the-art, open-source reference models on this dataset that we encourage others to
build on.

For future work, we plan to compare our CNN-based
approaches with an emerging family of models based on recurrent
architectures. We have not undertaken such a study because there do
not appear to be publicly-available reference implementations of such
models, and the lack of a common benchmark makes comparisons
difficult. The latter problem has been addressed,
and it would be interesting to see how recurrent neural
networks stack up against our approach.


\end{document}